\documentclass{article} 
\usepackage{iclr2027_conference,times}

\usepackage{amsmath,amsfonts,bm}

\def\eqref#1{equation~\ref{#1}}

\def\1{\bm{1}}

\DeclareMathAlphabet{\mathsfit}{\encodingdefault}{\sfdefault}{m}{sl}
\SetMathAlphabet{\mathsfit}{bold}{\encodingdefault}{\sfdefault}{bx}{n}

\usepackage{hyperref}
\usepackage{url}
\usepackage{graphicx}
\usepackage{cleveref}
\usepackage{xspace}
\usepackage{booktabs}
\usepackage{multirow}
\usepackage{algorithm}
\usepackage{algpseudocode}

\newcommand{\mempp}{Mem++\xspace}
\newcommand{\memg}{Mem$^{g}$++\xspace}

\usepackage{xcolor}
\usepackage{fontawesome5}
\definecolor{linkblue}{RGB}{70,130,180}
\hypersetup{colorlinks=true, urlcolor=linkblue, linkcolor=black, citecolor=black}

\title{\centering\upshape Mem++: Non-Destructive Memory for Long-Term\\ Organizational LLM Agents}

\author{\makebox[\dimexpr\textwidth-2\tabcolsep\relax]{\begin{tabular}{c}
Ahmad Yehia$^{1}$, Aly O. Abdelkareem$^{2}$, Islam Ahmed$^{2}$, Hesham Omran$^{2}$,\\
Khaled Alashmouny$^{2}$, Christian Claudel$^{1}$, Abduallah Mohamed$^{2}$\\[4pt]
\normalfont $^{1}$The University of Texas at Austin \quad $^{2}$AIDAChip Inc.\\[4pt]
\normalfont \texttt{ahmad.yehia@utexas.edu} \quad \texttt{abdu@aidachip.com}%
\thanks{\texttt{christian.claudel@utexas.edu} -- \texttt{\{aly, islam, hesham, khaled\}@aidachip.com}.}
\end{tabular}}}

\iclrfinalcopy 
\begin{document}

\maketitle
\lhead{Mem++: Long-Term Organizational Memory}

\begin{abstract}
Large Language Model (LLM) agents now take part in organizational work, where many authors record decisions across documents over months.
Because a revised decision arrives as a new document rather than an edit, answering a question requires knowing which version held at a given time.
However, most memory systems compress the record at write time.
By distilling each document into facts, notes or graph edges, these methods fix what can be answered before any question is asked.
To address this, we propose \mempp{}, a non-destructive memory framework shifting from write-time distillation to read-time selection.
\mempp{} stores every document whole with its date and author, and it calls no generative model at write time.
At read time, it retrieves only documents dated up to the time a question asks about and fuses lexical and semantic rankings.
Unlike systems that overwrite older versions, \mempp{} keeps them and leaves the choice to the answering model.
Evaluations on the organizational benchmark OrgMemBench demonstrate that \mempp{} surpasses the strongest memory system baseline by 8.0 to 13.1 points across two answering models.
With gpt-4.1-mini, it also achieves the best overall score, 2.6 points above RAG.
In addition, \mempp{} achieves the best average LLM-judge score on LoCoMo and ranks second on LongMemEval$_\text{S}$, behind only its entity-graph variant.
Code for benchmark evaluation is available at \href{https://github.com/AIDAChip-Inc/mem-plus-plus}{https://github.com/AIDAChip-Inc/mem-plus-plus}

\end{abstract}

\section{Introduction}

Large Language Model (LLM) agents now take part in organizational work, from daily tasks to decisions that span many teams over months \citep{kirchdorfer2026organizational}. Such critical work is only possible if the agent can correctly recall what was already decided \citep{zeng2025routine}. In an organization, the record of past decisions is shared; organizational memory keeps the history that informs present decisions \citep{walsh1991organizational}, and it is written by humans and agents alike.
To have a deeper understanding of the context and intent of the organization, a shared memory that mixes agents, humans, and compound their knowledge becomes a necessity.

\begin{figure}[t]
\begin{center}
\includegraphics[width=0.85\linewidth]{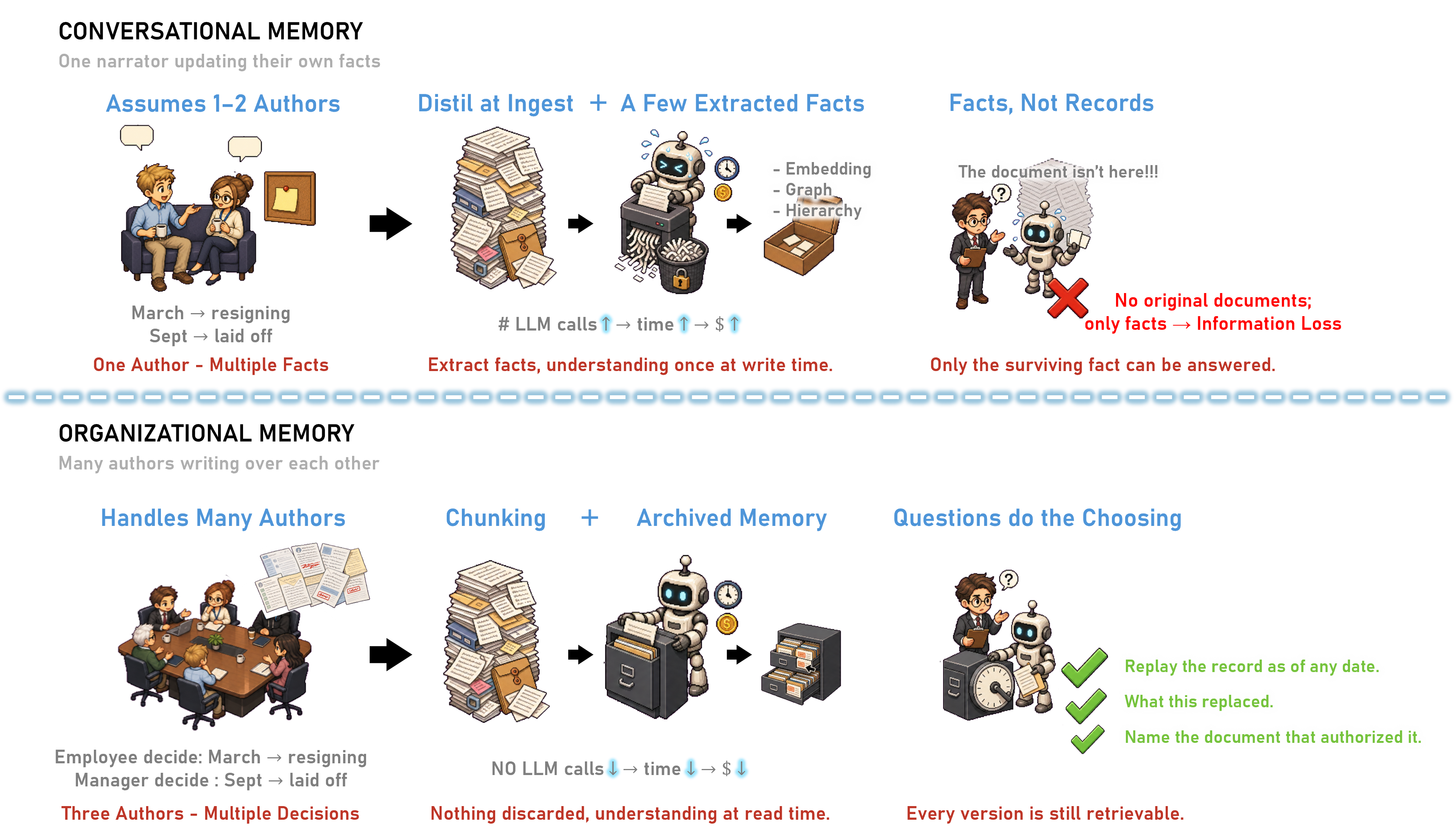}
\end{center}
\caption{\textbf{Conversational vs.\ organizational memory.}
\textbf{Top:} a single narrator restates their own facts, and LLM extraction at ingest keeps only the surviving fact.
\textbf{Bottom:} different authors in an organization write the same fact, and an earlier decision may still hold.
}
\label{fig:conv-vs-org}
\end{figure}

Organizational memory differs from the conversational, single-agent memory that existing systems were built for (\Cref{fig:conv-vs-org}).
Conversational memory has a single dialogue whose statements accumulate rather than compete, and a later statement is therefore read as an update of an earlier one.
Organizational memory, in contrast, is written by many people independently of one another across email, tickets, meeting notes and shared documents \citep{gardner2026orgmembench}.
Moreover, a decision that changes an earlier one is written as a new document rather than as an edit to the old one.
For example, a policy set in March, revised in June and reversed in September exists as three documents, and all three remain because nobody goes back to remove them.
Consequently, answering a question against this record takes more than finding the most recent version.
Conversational memory struggles here, because they treat every statement as coming from the same authority.
For example, Mem0\citep{chhikara2025mem0} resolves the conflict by overwriting, so the approved decision is replaced by the proposal and can no longer be retrieved afterwards.
In addition, Zep~\citep{rasmussen2025zep} keeps both statements, because its Graphiti engine stamps every edge with the valid period, but it stores extracted facts rather than the documents.

Existing memory systems resolve one trade-off at ingest, between keeping the record and structuring it.
Traditional retrieval-based systems keep the raw text and search it by similarity \citep{packer2023memgpt}, and nothing is discarded, but nothing is organized either.
A query, therefore, cannot distinguish a superseded value from one still in force.
This is how the memory stream of generative agents works, appending each observation in its original form and retrieving it later by relevance and recency \citep{park2023generative}.
Structured systems make the opposite choice. They record how statements relate to one another at ingest, and the raw text does not survive that step.
For instance, Mem0 stores a set of facts \citep{chhikara2025mem0}, Zep a graph of dated edges \citep{rasmussen2025zep}, and HippoRAG and GraphRAG an entity graph extracted from the corpus \citep{gutierrez2024hipporag,edge2024local}, and in none of them does a later query reach the document itself.
A memo that changes a decision is reduced to a relation between an entity and a value.
Its reason, condition and approval are not entity-shaped, and they never enter the store.
What will be answerable is therefore fixed before any question has been asked.
As we show in \Cref{sec:exp}, this explains the sharp drop of conversational memory systems on OrgMemBench, a benchmark of multi-author temporal, contradiction and justification-chain questions that define organizational memory.

To address these limitations we introduce \textbf{Mem++}, an organizational memory architecture that moves the decision about what matters from write time to read time.
(i) Every artifact is stored whole and remains the unit of retrieval.
(ii) Structure can be added as an optional layer over the retained documents rather than as a replacement for them.
A \mempp pass groups near-duplicates and marks older conflicting versions as superseded without deleting them, and we evaluate it as an ablation.
(iii) Selection is deferred until a question arrives, and it therefore ranges over the full record rather than a pruned set of facts.
What distinguishes \mempp{} is therefore not that the record is kept, but that each document stays whole with its date and the choice between versions is made at read time.
Our contributions are as follows:

\begin{itemize}

\item \textbf{\mempp{}, a non-destructive organizational memory.} We introduce a memory that stores each document whole with its date and author, calls no generative model at write time, and defers the choice between versions to read time.

\item \textbf{Evaluation.} We evaluate \mempp{} against publicly available memory systems on the recently released organizational benchmark OrgMemBench \citep{gardner2026orgmembench} and on the conversational benchmarks LoCoMo and LongMemEval$_\text{S}$, where it outperforms every memory baseline on OrgMemBench and remains competitive on the conversational benchmarks.

\item \textbf{Ablations.} We measure the contribution of each retrieval leg, toggle consolidation and extracted-fact cards, and compare with the entity graph of \memg{}.

\end{itemize}

\section{Related Work}

\subsection{Conversational Memory Architectures for LLM Agents}

Agent memories give an LLM persistent state beyond its context window, differing in how they carry information across sessions.
MemGPT \citep{packer2023memgpt} and the generative agents memory \citep{park2023generative} hold the record in full and leave every judgement to retrieval.
Details survives, but nothing marks how one statement stands to another, so a query cannot separate a replaced value from one still activated.
Retrieval-augmented generation (RAG) grounds answers in an external corpus by chunking documents, embedding them, and returning the nearest chunks at query time \citep{lewis2020retrieval,karpukhin2020dense}.
However, the words survive while the document does not, since chunking separates a passage from the rest of its text and from any later version of it.
Most recent approaches instead compress at ingest, gaining structure at the cost of the record.
Mem0 distils each document into facts and overwrites them on conflict \citep{chhikara2025mem0}.
Similarly, A-Mem, Nemori, LightMem and LiCoMemory rewrite interactions into episodes through clustering, summarization or note-taking \citep{xu2026mem,nan2025nemori,fang2026lightmem,huang2026licomemory}.
In addition, G-Memory organizes these episodes hierarchically \citep{zhang2026gmemory}, PREMem reasons as it writes \citep{kim2025premem}, and MemOS exposes memory as a managed layer \citep{li2025memos}.
Each gains a compact store but loses the source and the links between its statements, and what can be answered is fixed at ingest.
In contrast, \mempp stores every document whole at write time and marks superseded statements with a date instead of deleting them.
This approach keeps every version available and leaves the choice between them to read time, and conflicting statements from different authors therefore remain answerable.

\subsection{Long-term Temporal Memory Systems}
Temporal questions ask what held at a given time or what a newer statement replaced, and conversational benchmarks such as LongMemEval and LoCoMo test them through knowledge-update and temporal-reasoning categories \citep{wu2024longmemeval,maharana2024evaluating}.
Existing systems handle such changes in two ways.
The first discards or alters the older value.
For instance, Mem0 overwrites an older fact once a newer one arrives \citep{chhikara2025mem0}, A-Mem revises related notes in place when a new one is added \citep{xu2026mem}, and MemoryBank lets older memories fade along a forgetting curve \citep{zhong2024memorybank}.
The second keeps the older value with a date.
Zep and LongMemEval attach time intervals or timestamps to stored facts so that queries can be bounded by date \citep{rasmussen2025zep,wu2024longmemeval}.
In conversational memory, both mechanisms treat a change as one narrator updating their own facts, and the newest value is therefore assumed to be correct while the older one is removed or kept only as history.
In an organization, however, changes arrive as new documents from different authors across different channels, and questions ask about past versions as often as about the current state \citep{gardner2026orgmembench}.
\mempp addresses this by keeping every version as a stored document, marking the replaced one with the date it stopped holding.
Then, each retrieval is bounded by the date in the question, which lets one store answer what holds now, what held before, and what changed.
In other terms, it reconstructs the life journey of a fact and how it changed. This is useful for organization to track pivot points in decisions making, allowing the agents and humans to revisit these later.

\section{Method}
\label{sec:method}

\begin{figure}[t]
\centering
\includegraphics[width=0.85\textwidth]{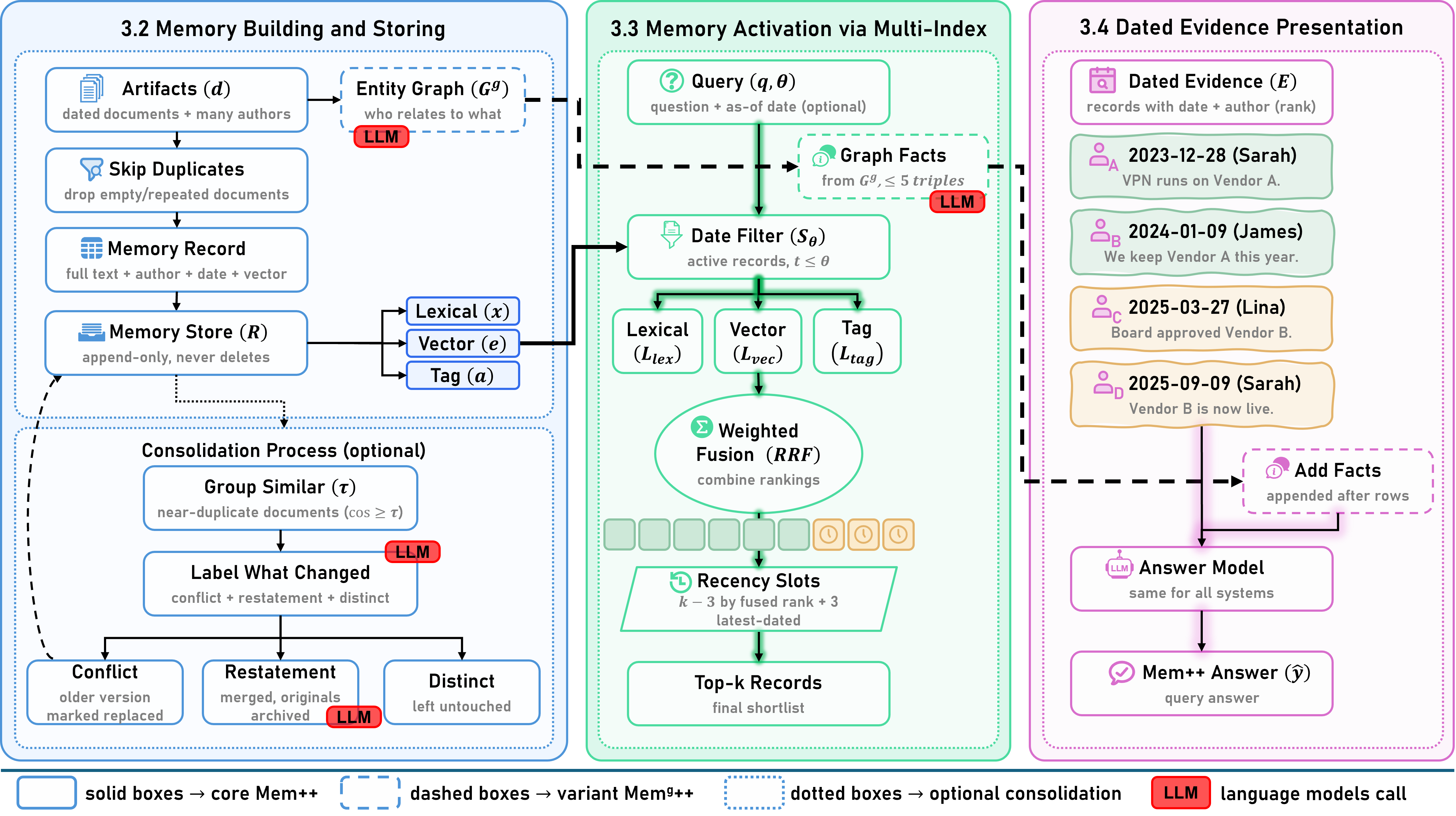}
\caption{Overview of \mempp framework.
\textbf{(3.2)}~Each document is stored whole with its author, date and sentence embedding, and it is indexed three ways.
\textbf{(3.3)}~Records valid at the as-of date~$\theta$ are retrieved from the three indexes and fused with weighted RRF, and three of the $k$ slots are reserved for the latest-dated matches.
\textbf{(3.4)}~The top-$k$ records are passed with their dates and authors to a fixed answering model.
\memg{} graph triples carry no date.}
\label{fig:architecture}
\end{figure}

Organizations keep records to preserve their decisions over time~\citep{walsh1991organizational}.
Unlike a conversation, this record is a stream of dated documents from many authors. 
A useful memory must therefore tell what the current value is, what the value was at a past date, and what a change replaced and who made it.
To this end, \mempp{} stores each new document whole, never rewrites or deletes one to resolve a conflict, and leaves the choice between versions to the answering model.
\Cref{fig:architecture} shows its three stages, namely a write path that stores documents, an optional consolidation pass that links their versions (\Cref{app:method}), and a read path that answers a question against the store.
\Cref{alg:mempp} in \Cref{app:method} gives the pseudocode.

\subsection{Architecture}
\label{subsec:arch}
Formally, \mempp{} consists of one store and two operators,
\begin{equation}
  \text{\mempp} = \langle \mathcal{R},\ \mathcal{C},\ \Phi \rangle ,
  \label{eq:mempp}
\end{equation}
where $\mathcal{R}$ is the memory store filled by the write path, $\mathcal{C}$ is an optional consolidation operator that relates versions, and $\Phi$ is the read operator that retrieves evidence for a question.
Each stored document becomes one record, or row,
\begin{equation}
  r = \langle x,\ a,\ t,\ \mathbf{e},\ \sigma \rangle ,
  \label{eq:row}
\end{equation}
where $x$ is its full text, $a$ is its author, $t$ is its event date, $\mathbf{e}$ is an embedding of $x$, and $\sigma$ is its state, which records whether the row is active and which later row, if any, superseded it.
Only $\sigma$ may change after a row is written, and no row is ever deleted.
Hence, the store $\mathcal{R}=\{r_1,\dots,r_N\}$ only grows, and every earlier version remains in the store with its original text.
The consolidation operator $\mathcal{C}$ groups near-duplicate rows and marks older conflicting versions as superseded in $\sigma$.
It is described in \Cref{app:method}, disabled in all headline runs and evaluated in the ablations.

\subsection{Memory Building and Storing}
\label{subsec:write}

Unlike approaches that segment the stream into windows or rewrite it into extracted units~\citep{chhikara2025mem0}, the write path stores each document once and as a whole, so no segment boundary splits a document and no model guesses at write time what a later question may need.

\paragraph{Whole-Document Ingestion.}
For each document, the write path creates one row $r$ of the form in Eq.~\eqref{eq:row}, with embedding $\mathbf{e}=h(x)$ from a sentence encoder $h(\cdot)$ (\Cref{tab:settings}), and sets $\sigma$ to mark the row as current and active.
The text is indexed both in a full-text index for lexical search and through $\mathbf{e}$ for vector search, and no generative model is called on this path.
Documents longer than the encoder limit are truncated only in $\mathbf{e}$, while $x$ and the lexical index keep the full text.

\paragraph{Provenance and Time Anchoring.}
The author is recorded as a tag wherever the source names one, and this tag feeds the tag index.
On OrgMemBench the benchmark loader passes no author field, so this tag stays empty for every system, although names appear inside the text.
The event date is stored in $t$ whenever the source provides one, as OrgMemBench documents do \citep{gardner2026orgmembench}.
On the conversational benchmarks, each turn is stored with its session date in $t$, and the date and speaker also lead its text.

\paragraph{Append-Only Incremental Updates.}
New documents are added without touching the rows already stored.
The write path skips a document whose text is empty or exactly repeats an active row in the same author scope, so re-ingesting the same stream adds no duplicate current rows.
Adding a document thus costs one embedding, and no older row is reprocessed.

\subsection{Retrieval via Multi-Index Fusion}
\label{subsec:read}

Given a question $q$ and an optional as-of date $\theta$, the read path selects the $k$ rows that the answering model will see:

\begin{equation}
  E = \Phi(q, \theta \mid \mathcal{R}) = \mathrm{TopK}\big(\mathrm{RRF}(L^{\theta}_{\mathrm{lex}},\ L^{\theta}_{\mathrm{tag}},\ L^{\theta}_{\mathrm{vec}})\big) ,
\end{equation}
where $L^{\theta}_\ell$ is the ranked list that index $\ell \in \{\mathrm{lex}, \mathrm{tag}, \mathrm{vec}\}$ returns for $q$ within the temporal scope defined below, and $\mathrm{TopK}$ keeps $k$ rows as described below.
Because questions range from exact identifiers to loose paraphrases, \mempp{} queries three indexes over one shared scope and fuses their rankings without a language model call.

\paragraph{Temporal Scoping.} All three indexes draw from the same base set of rows,
\begin{equation}
  \mathcal{S}_\theta = \{\, r \in \mathcal{R} \mid \mathrm{active}(\sigma_r),\ t_r \le \theta \,\} ,
\end{equation}
where the time condition is dropped when the question carries no as-of date.
Because this condition is applied before any index ranks a row, all $k$ slots go to eligible rows, whereas filtering after ranking could leave fewer than $k$.
Rows without an event date fail the time condition, so an as-of question can only reach dated rows.

\paragraph{Lexical, Tag, and Semantic Pathways.} Within $\mathcal{S}_\theta$, each index returns its own ranked list of at most $C$ candidates. The lexical index matches the content words of $q$ against the full-text, and ranks matches by cover density, recovering exact identifiers and an embedding may blur. The tag index matches $q$, against author tags; since the reported configurations carry only author labels, it rarely fires and is evaluated as an ablation. The semantic index ranks rows by the cosine similarity between $h(q)$ and $\mathbf{e}$, recovering paraphrases that share no words with the question.

\paragraph{Weighted Rank Fusion with Recency Reserve.} The three lists are merged with weighted reciprocal rank fusion~\citep{cormack2009reciprocal}, which scores a row by
\begin{equation}
  \mathrm{RRF}(r) = \sum_{\ell \in \{\mathrm{lex},\,\mathrm{tag},\,\mathrm{vec}\}} \frac{w_\ell}{\gamma + \mathrm{rank}_\ell(r)} ,
\end{equation}
where $w_\ell$ is the weight of index $\ell$, $\mathrm{rank}_\ell(r) \in \{0, 1, \dots\}$ is the position of $r$ in $L^{\theta}_\ell$, $\gamma$ is a smoothing constant, and a row absent from a list receives nothing from it.
Because fusion uses ranks rather than raw scores, the three pathways need no score normalization. Ties are broken by ingest recency. Of the $k$ returned rows, $k-3$ follow this order, and the other three go to the remaining matched rows with the latest event date, falling back to ingest time where no event date exists. If fewer than three matched rows remain, the reserve is filled with the most recently ingested unmatched rows. 
Read operator $\Phi$ does not read the supersession links in $\sigma$, so a superseded row competes like any other active row.

\paragraph{Graph Retrieval (\memg).} At ingest, \memg extracts entity-relation triples from each document with a language model, and these triples form the graph.
At query time,  one model call extracts the entities named in, and the graph  is searched from those entities to collect related triples.
These graph results are appended after the fused rows, at most five per question, so \memg sees more context than the systems it is compared against.
Triples carry no date and the as-of condition is not applied to them, so on a dated question the graph can surface relations from artifacts later than $\theta$.

\paragraph{Graph Retrieval (\memg{}).}
\memg{} adds to \mempp{} the entity-relation graph of Mem0$^g$~\citep{chhikara2025mem0}.
At write time, two gpt-4o-mini calls extract entity-relation triples from each document, and these triples form the graph $\mathcal{K}$.
At query time, one model call extracts the entities $V_q$ named in $q$, and $\mathcal{K}$ is searched from those entities to collect related triples.
These triples are appended after the fused rows, at most five per question. This is why on the conversational benchmarks, \memg{} sees more context than the systems it is compared against.
The triples carry no date and the as-of condition is not applied to them, so on a dated question the graph can surface relations drawn from documents later than $\theta$.

\subsection{Evidence Presentation} Each row is rendered in rank order as its original text preceded by its event date and, where the source records one, an author tag, so that the model can compare versions of a fact by when they were written. On OrgMemBench the header is built from $t$ and that tag, although no artifact on the reported tier names an author, so every row's header carries the same placeholder author alongside the date. On the conversational suites the date and speaker already lead the stored text, so each row is shown as stored, preceded only by its rank number. This design keeps \mempp a memory rather than a reasoner, so it adds no inference step of its own at answer time.

\section{Experiments}
\label{sec:exp}

\subsection{Experimental Setup}
\label{subsec:exp_setup}

\begin{figure}[t]
\centering
\includegraphics[width=0.85\textwidth]{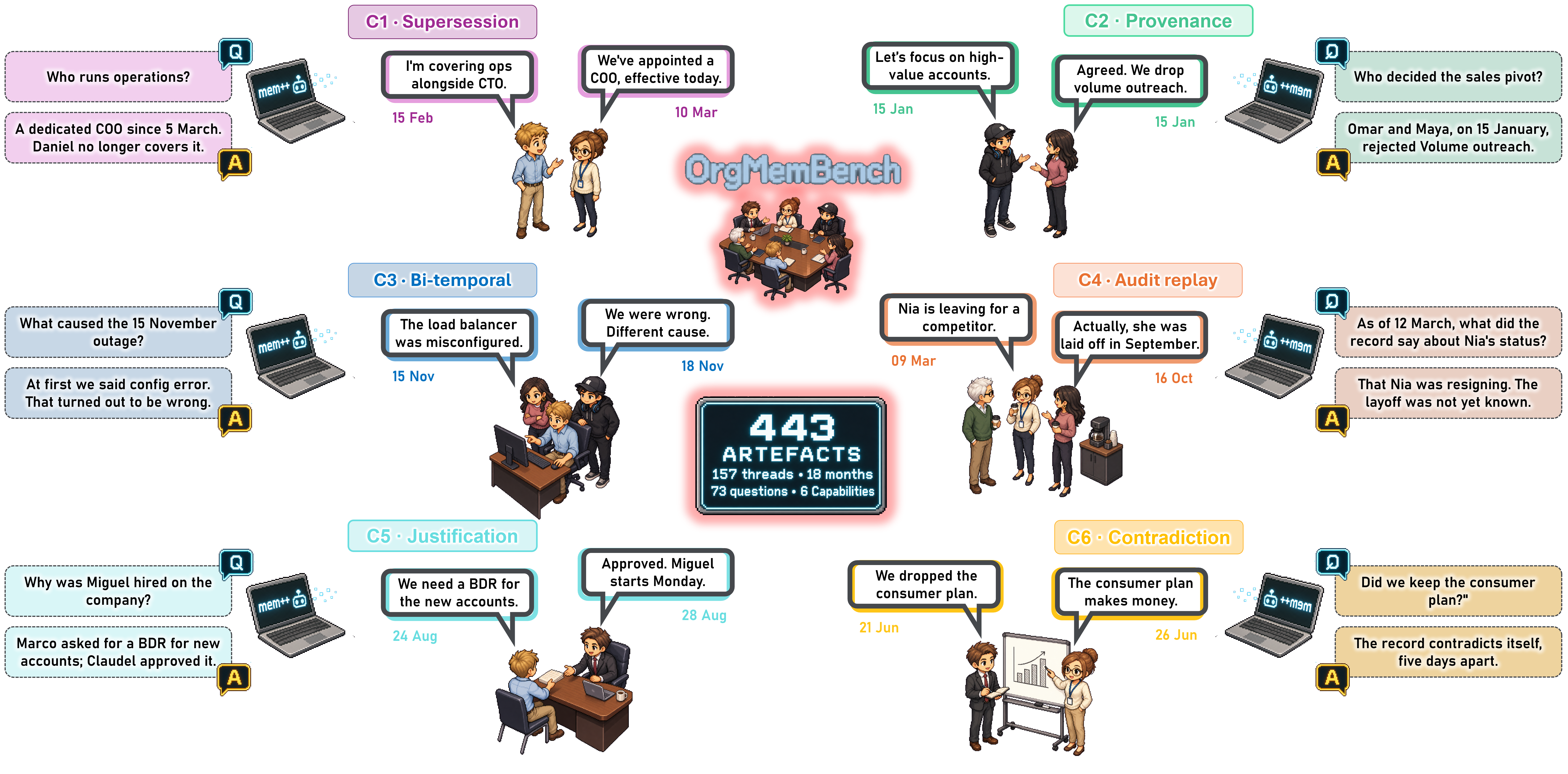}
\caption{The six capabilities of OrgMemBench~\citep{gardner2026orgmembench}, with questions and answers examples.
The tier we evaluated holds 443 artifacts across 157 threads for 18 months, with 73 questions.}
\label{fig:orgmembench}
\end{figure}

\paragraph{Datasets.}
We evaluate \mempp on three benchmarks, one organizational and two conversational, which together cover the settings it is \mempp designed for.
\textbf{OrgMemBench}~\citep{gardner2026orgmembench} is our primary benchmark, a synthetic bi-temporal corpus of one fictional organization that we use unmodified.
We run its medium tier, which pairs 443 dated artifacts, organized into 157 threads over 18 months where later decisions revise earlier ones, with 73 questions over six capabilities that range from supersession to contradiction (Figure~\ref{fig:orgmembench}).
The question count is small, yet each answer is graded against a rubric rather than marked right or wrong, giving 247 graded facets and a score in $[0,1]$ that reflects partial coverage of the expected evidence.
\textbf{LoCoMo}~\citep{maharana2024evaluating} provides 10 multi-session dialogues of about 24K tokens each and 1,540 questions across four reasoning categories, and it is the most widely used for memories.
\textbf{LongMemEval\textsubscript{S}}~\citep{wu2024longmemeval} keeps a similar question format while spreading each of its 500 questions over roughly 48 sessions and 494 turns, which tests if a memory still holds when the history grows.

\paragraph{Baselines.}
We benchmark against seven methods that range from no external memory to richly structured ones, namely \textbf{Full Context} (entire history in the prompt), \textbf{RAG-4096} \citep[4096-token chunks for dense retrieval]{lewis2020retrieval}, \textbf{Mem0} \citep[fact extraction with update and delete rules]{chhikara2025mem0}, \textbf{Zep} \citep[temporal knowledge graph with dated edges]{rasmussen2025zep}, \textbf{A-Mem} \citep[structured notes with evolving links]{xu2026mem}, \textbf{LangMem} \citep[automatic memory extraction across sessions]{langmem2025}, and \textbf{gbrain} \citep[entity graph over organizational documents]{gardner2026orgmembench}.
Full Context, RAG-4096, Mem0 and Zep serve as general-purpose baselines and run on all three benchmarks, whereas A-Mem and LangMem, which designed for conversational memory, run only on LoCoMo and LongMemEval\textsubscript{S}, and gbrain, which was released with OrgMemBench, runs only there.

\paragraph{Evaluation Metrics.}
On all three datasets, our primary evaluation metric is the \textbf{LLM-judge score} (abbreviated as \textbf{LLM} for simplicity), using gpt-4o-mini as the judges.
On LoCoMo and LongMemEval\textsubscript{S}, we additionally report \textbf{F1} and \textbf{BLEU-1}.
For OrgMemBench, the judges grade against the benchmark's rubric.
For LongMemEval\textsubscript{S}, prompts are adapted to its task-specific question-answering format, following Zep~\citep{rasmussen2025zep}.
These metrics are accuracy metrics of different standard, and are scaled to the 0--100 range, with higher values indicating better performance and 100 denoting a perfect score.

\paragraph{Implementation Details.}
To ensure fair comparison, Mem0 and Zep retrieve memory contexts, which are then fed to gpt-4o-mini, gpt-4.1-mini for answer generation.
All other methods, including \mempp, employ the same models for answer generation, while \memg uses gpt-4o-mini as its internal graph backbone.
All retrieval-based baselines employ the powerful text-embedding-3-small model, whereas \mempp uses the lightweight all-MiniLM-L6-v2 model. The lexical index matches query terms disjunctively by default and conjunctively on OrgMemBench.
Key hyperparameters are the fusion constant $\gamma$ and the pathway weights $w_\ell$, while the consolidation threshold $\tau$ applies only when consolidation is enabled.
For retrieval count, we fix $k=50$ in main experiments, while $k$ varies from 10 to 100 in the ablation study, as shown in the Appendix.

\begin{table}[t]
\centering
\caption{Performance on OrgMemBench \citep{gardner2026orgmembench} categorized by question type. Overall is weighted by the number of questions in each category and reported with its standard deviation across runs. Bold indicates the best performance. Underline indicates the second best performance.}
\label{tab:orgmem}
\resizebox{\linewidth}{!}{%
\begin{tabular}{llccccccc}
\toprule
& Method & Supersession & Decision Provenance & Bi-temporal & Audit Replay & Justification Chain & Contradiction & Overall \\
\midrule
\multirow{8}{*}{\rotatebox[origin=c]{90}{\textbf{gpt-4.1-mini}}}
& Full Context &  0.0 & 27.5 &  0.0 & 33.0 & 22.7 &  8.3 & 17.8$\pm$0.1 \\
\cmidrule{2-9}
& RAG      & 66.1 & \textbf{79.7} & \textbf{66.7} & 40.9 & 19.6 & \textbf{75.4} & 55.0$\pm$0.3 \\
& Zep      & 31.4 & 42.5 & 23.0 & 76.8 & 20.7 & 43.3 & 41.0$\pm$0.2 \\
& Mem0     & 40.8 & 55.0 & 50.0 & 25.1 & 14.4 & \underline{52.8} & 36.9$\pm$0.0 \\
& A-Mem    & 58.3 & 63.3 & 43.6 & 36.5 & 20.7 & 43.5 & 44.5$\pm$0.3 \\
& gbrain   & 54.8 & 60.7 & 30.3 & 48.5 & \textbf{22.7} & 22.9 & 43.2$\pm$0.1 \\
\cmidrule{2-9}
& \mempp & \textbf{67.7} & \underline{65.9} & 50.0 & \textbf{83.0} & \underline{21.5} & 47.2 & \textbf{57.6}$\pm$0.1 \\
& \memg  & \underline{67.2} & 63.9 & \underline{50.6} & \underline{79.7} & 19.8 & 44.4 & \underline{55.9}$\pm$0.1 \\
\midrule
\multirow{8}{*}{\rotatebox[origin=c]{90}{\textbf{gpt-4o-mini}}}
& Full Context &  2.5 & 21.1 &  0.0 & 35.7 & 17.0 & 13.9 & 16.8$\pm$0.1 \\
\cmidrule{2-9}
& RAG      & \underline{56.7} & 47.8 & 30.2 & 55.1 & \underline{10.4} & \textbf{67.2} & 43.4$\pm$0.2 \\
& Zep      & 27.5 & 38.3 & \textbf{34.9} & \textbf{74.9} & 7.4 & 29.6 & 36.2$\pm$0.3 \\
& Mem0     & 32.3 & 47.2 & \underline{33.3} & 32.4 & \textbf{11.8} & \underline{62.8} & 33.8$\pm$0.1 \\
& A-Mem    & 42.8 & \textbf{60.0} & 28.6 & 21.3 & 4.4 & 41.7 & 32.6$\pm$0.3 \\
& gbrain   & 47.8 & 46.1 & 25.9 & 46.6 & 7.7 & 21.9 & 34.7$\pm$0.2 \\
\cmidrule{2-9}
& \mempp & \textbf{57.0} & \underline{49.2} & \underline{33.3} & 70.7 & 8.9 & 39.8 & \underline{44.2}$\pm$0.1 \\
& \memg  & \underline{56.7} & 48.6 & 32.1 & \underline{71.9} & 8.8 & 37.7 & \textbf{44.4}$\pm$0.1 \\
\bottomrule
\end{tabular}%
}
\end{table}

\begin{table}[t]
\centering
\caption{Performance on LoCoMo \citep{maharana2024evaluating} categorized by type. Bold, underline indicates best and second best performance. The baseline performance comes from \citet{nan2025nemori}.}
\label{tab:locomo}
\resizebox{\linewidth}{!}{%
\begin{tabular}{llccccccccccccccc}
\toprule
& \multirow{2}{*}{Method}
& \multicolumn{3}{c}{Temporal Reasoning}
& \multicolumn{3}{c}{Open Domain}
& \multicolumn{3}{c}{Multi-Hop}
& \multicolumn{3}{c}{Single-Hop}
& \multicolumn{3}{c}{Average} \\
\cmidrule(lr){3-5}\cmidrule(lr){6-8}\cmidrule(lr){9-11}\cmidrule(lr){12-14}\cmidrule(lr){15-17}
& & $\uparrow$LLM & $\uparrow$F1 & $\uparrow$BLEU
  & $\uparrow$LLM & $\uparrow$F1 & $\uparrow$BLEU
  & $\uparrow$LLM & $\uparrow$F1 & $\uparrow$BLEU
  & $\uparrow$LLM & $\uparrow$F1 & $\uparrow$BLEU
  & $\uparrow$LLM & $\uparrow$F1 & $\uparrow$BLEU \\
\midrule
\multirow{9}{*}{\rotatebox[origin=c]{90}{\textbf{gpt-4.1-mini}}}
& Full Context & 74.2 & 47.5 & 40.0 & 56.6 & 28.4 & 22.2 & 77.2 & 44.2 & 33.7 & 86.9 & 61.4 & 53.4 & 80.6 & 53.3 & 45.0 \\
\cmidrule{2-17}
& RAG-2048 & 66.8 & \underline{50.3} & \underline{40.1} & 48.6 & \underline{28.3} & 21.7 & 67.4 & 39.1 & 28.9 & 82.8 & 57.7 & 48.1 & 74.5 & 50.9 & 41.3 \\
& RAG-4096 & 27.4 & 22.3 & 19.1 & 28.8 & 17.9 & 13.9 & 31.7 & 20.1 & 12.8 & 35.9 & 25.8 & 22.0 & 32.9 & 23.5 & 19.2 \\
& Zep      & 60.2 & 23.9 & 20.0 & 43.8 & 24.2 & 19.3 & 53.7 & 30.5 & 20.4 & 66.9 & 45.5 & 40.0 & 61.6 & 36.9 & 30.9 \\
& Mem0     & 56.9 & 39.2 & 33.2 & 47.9 & 23.7 & 17.7 & 68.2 & 40.1 & 30.3 & 71.4 & 48.6 & 42.0 & 66.3 & 43.5 & 36.5 \\
& A-Mem    & 66.7 & 40.3 & 33.7 & 37.5 & 13.4 & 12.7 & 55.7 & 30.4 & 20.0 & 64.0 & 45.0 & 39.8 & 61.4 & 39.4 & 33.2 \\
& Nemori   & 77.6 & \textbf{57.7} & \textbf{50.2} & \underline{51.0} & 25.8 & 19.3 & \textbf{75.1} & \underline{41.7} & \underline{31.9} & 84.9 & 58.8 & 51.5 & 79.4 & \textbf{53.4} & \textbf{45.6} \\
\cmidrule{2-17}
& \mempp & \textbf{81.4} & 41.7 & 33.5 & \textbf{53.0} & \textbf{29.0} & \underline{22.1} & \underline{74.8} & \textbf{41.9} & \textbf{32.0} & \textbf{87.0} & \textbf{62.1} & \underline{54.8} & \textbf{81.5} & \underline{52.1} & 44.2 \\
& \memg  & \underline{81.1} & 41.8 & 33.6 & 50.5 & 27.9 & \textbf{22.8} & 71.8 & 41.2 & \textbf{32.0} & \underline{86.4} & \underline{62.0} & \textbf{55.0} & \underline{80.4} & 51.9 & \underline{44.3} \\
\midrule
\multirow{9}{*}{\rotatebox[origin=c]{90}{\textbf{gpt-4o-mini}}}
& Full Context & 56.2 & 44.1 & 36.1 & 48.6 & 24.5 & 17.2 & 66.8 & 35.4 & 26.1 & 83.0 & 53.1 & 44.7 & 72.3 & 46.2 & 37.8 \\
\cmidrule{2-17}
& RAG-2048  & 62.6 & 47.5 & 38.2 & 45.2 & 23.4 & 18.1 & 63.0 & 35.7 & 24.6 & 78.5 & 53.6 & 43.0 & 70.3 & 47.2 & 37.1 \\
& RAG-4096 & 22.8 & 18.4 & 15.2 & 35.5 & 17.6 & 15.6 & 31.1 & 18.6 & 11.7 & 33.0 & 23.9 & 19.2 & 30.7 & 21.4 & 16.8 \\
& Zep      & 58.9 & 44.8 & 38.1 & 39.6 & 22.9 & 15.7 & 50.5 & 27.5 & 19.3 & 63.2 & 39.7 & 33.7 & 58.5 & 37.5 & 30.9 \\
& Mem0     & 50.4 & 44.4 & 37.6 & 40.6 & \textbf{27.1} & \textbf{19.4} & 60.3 & 34.3 & 25.2 & 68.1 & 44.4 & 37.7 & 61.3 & 41.5 & 34.2 \\
& A-Mem    & 54.2 & 38.1 & 33.8 & 22.9 &  9.0 &  8.6 & 43.6 & 24.0 & 18.8 & 58.2 & 35.6 & 29.2 & 52.5 & 32.4 & 27.0 \\
& Nemori   & 71.0 & \underline{56.7} & \textbf{46.6} & 44.8 & 20.8 & 15.1 & 65.3 & 36.5 & \underline{25.6} & 82.1 & 54.4 & 43.2 & 74.4 & 49.5 & 38.5 \\
\cmidrule{2-17}
& \mempp & \textbf{76.9} & \textbf{56.9} & \underline{46.5} & \textbf{47.3} & 24.2 & 17.8 & \textbf{68.5} & \textbf{39.0} & \textbf{27.5} & \underline{84.0} & \textbf{59.7} & \textbf{47.9} & \textbf{77.4} & \textbf{53.1} & \textbf{42.0} \\
& \memg  & \underline{75.6} & 55.3 & 44.5 & \underline{46.2} & \underline{24.6} & \underline{18.8} & \underline{65.8} & \underline{37.2} & 25.3 & \textbf{84.1} & \underline{58.9} & \underline{47.3} & \underline{76.7} & \underline{52.1} & \underline{41.0} \\
\bottomrule
\end{tabular}%
}
\end{table}

\begin{table}[t]
\centering
\caption{Performance on LongMemEval$_\text{S}$ \citep{wu2024longmemeval} by question type. LLM-judge accuracy is reported. Bold indicates the best performance. Underline indicates the second best performance.}
\label{tab:longmemeval}
\small
\begin{tabular}{llccccc}
\toprule
 & Question Type & Full Context & Zep & Nemori & \mempp & \memg \\
\midrule
\multirow{7}{*}{\rotatebox{90}{\textbf{gpt-4o-mini}}}
 & Single-session preference & 6.7 & 20.0 & \underline{46.7} & \underline{46.7} & \textbf{50.0} \\
 & Single-session assistant & 89.3 & 80.4 & 83.9 & \underline{94.6} & \textbf{96.4} \\
 & Temporal reasoning & 42.1 & \textbf{62.4} & \underline{61.7} & 56.7 & 56.7 \\
 & Multi-session & 38.3 & 57.9 & 51.1 & \textbf{63.4} & \underline{62.0} \\
 & Knowledge update & 78.2 & 83.3 & 61.5 & \underline{84.3} & \textbf{85.2} \\
 & Single-session user & 78.6 & \underline{92.9} & 88.6 & \textbf{98.4} & \textbf{98.4} \\
\cmidrule{2-7}
 & Average & 55.0 & 68.0 & 64.2 & \underline{72.2} & \textbf{72.4} \\
\midrule
\multirow{7}{*}{\rotatebox{90}{\textbf{gpt-4.1-mini}}}
 & Single-session preference & 16.7 & 22.5 & \textbf{86.7} & 47.8 & \underline{53.3} \\
 & Single-session assistant & \textbf{98.2} & 83.1 & 92.9 & \underline{94.6} & \underline{94.6} \\
 & Temporal reasoning & 60.2 & 64.5 & \textbf{72.2} & 68.5 & \underline{69.0} \\
 & Multi-session & 51.1 & \textbf{57.6} & 55.6 & 56.1 & \underline{56.8} \\
 & Knowledge update & 76.9 & \underline{83.1} & 79.5 & \textbf{89.9} & \textbf{89.9} \\
 & Single-session user & 85.7 & \underline{96.3} & 90.0 & \textbf{100.0} & \textbf{100.0} \\
\cmidrule{2-7}
 & Average & 65.6 & 69.4 & 74.6 & \underline{74.7} & \textbf{75.3} \\
\bottomrule
\end{tabular}
\end{table}

\subsection{Main Results}
\label{sec:main-results}
Tables~\ref{tab:orgmem}, \ref{tab:locomo} and \ref{tab:longmemeval} report the results on OrgMemBench, LoCoMo and LongMemEval$_\text{S}$.
Regarding these tables, we highlight the following observations.

\textbf{Strong Performance on Organizational Memory.}
\mempp{} achieves the best overall score on OrgMemBench with gpt-4.1-mini, reaching 57.6 and surpassing the strongest baseline, RAG, by 2.6 points.
With gpt-4o-mini, \memg{} and \mempp{} rank first and second with 44.4 and 44.2, ahead of RAG at 43.4.
The memory systems fall further behind, and \mempp{} exceeds the best of them by 13.1 points over A-Mem with gpt-4.1-mini and by 8.0 points over Zep with gpt-4o-mini.
Full Context performs worst, scoring below 18 overall and near zero on Supersession and Bi-temporal questions with both answerers.
This suggests that placing the whole record in context does not by itself reveal which version holds.

\textbf{Supersession and Audit Replay.}
\mempp{} is the best method on Supersession with both answerers, scoring 67.7 and 57.0.
It also leads Audit Replay with gpt-4.1-mini at 83.0, which is 6.2 points above Zep, while \memg{} leads it with gpt-4o-mini at 71.9.
We attribute the gap over Mem0 and Zep to keeping every original document in the store with its date and author.
In contrast, Mem0 overwrites a fact on conflict, which likely explains its Supersession scores of 40.8 and 32.3.

\textbf{Competitive Performance on Conversational Memory.}
Although \mempp{} targets organizational records, it also achieves the best average LLM-judge score on LoCoMo with both answerers.
With gpt-4.1-mini, it reaches 81.5, surpassing Nemori by 2.1 points and Full Context by 0.9 points.
With gpt-4o-mini, it reaches 77.4, exceeding Nemori by 3.0 points and Full Context by 5.1 points.
The largest LLM-judge gain over Nemori is on Temporal Reasoning, where \mempp{} scores 81.4 and 76.9, which is 3.8 and 5.9 points higher.
We attribute this to the recency slots, which reserve part of the context for the latest-dated matching documents.

\textbf{Performance on Long Contexts.}
LongMemEval$_\text{S}$ conversations exceed 100K tokens, an order of magnitude more than LoCoMo \citep{nan2025nemori}.
On this benchmark, \memg{} achieves the best average with both answerers.
It reaches 72.4 with gpt-4o-mini, 4.4 points above Zep, and 75.3 with gpt-4.1-mini, 0.7 points above Nemori.
\mempp{} ranks second with 72.2 and 74.7, and both variants exceed Full Context by at least 9.1 points.
Moreover, the two variants hold the top two positions on Knowledge Update and Single-session User with both answerers.

\textbf{Where \mempp{} Trails.}
\mempp{} does not lead every category.
On OrgMemBench, RAG exceeds \mempp{} on Contradiction by 28.2 and 27.4 points and on Bi-temporal with gpt-4.1-mini by 16.7 points.
In addition, every method stays below 23 on Justification Chain.
On LongMemEval$_\text{S}$, \mempp{} trails on Temporal Reasoning with both answerers, and Nemori leads Single-session Preference with gpt-4.1-mini (86.7 vs.\ 47.8).
We attribute the preference gap to Nemori's distillation, which can infer preferences that are rarely seen.
On LoCoMo, \mempp{} also trails Nemori on Temporal F1 with gpt-4.1-mini (41.7 vs.\ 57.7).
However, the two methods are level with gpt-4o-mini (56.9 vs.\ 56.7), suggesting that the gap stems from answers phrasing rather than retrieval.

\textbf{Limited Gain from the Graph.}
The graph in \memg{} changes the overall score by at most 1.1 points in Tables~\ref{tab:orgmem}--\ref{tab:longmemeval}.
Since the graph adds two LLM calls per document at write and one per question at read, we consider \mempp{} the better trade-off between accuracy and cost.

\subsection{Ablation Study}
\label{sec:ablation}
Table~\ref{tab:ablation} reports the ablation with claude-sonnet-4-6 as the answerer with the same gpt-4o-mini judge.

\begin{table}[t]
\centering
\providecommand{\dlt}[1]{\,{\scriptsize(#1)}}
\caption{Ablation with claude-sonnet-4-6 as the answerer.
Parentheses give the change from \mempp{}.}
\label{tab:ablation}
\small
\begin{tabular}{lccc}
\toprule
Variant & OrgMemBench & LoCoMo & LongMemEval$_\text{S}$ \\
\midrule
\mempp{} & 57.0 & 85.7 & 87.9 \\
\midrule
w/o vector leg & 21.3\dlt{$-$35.7} & 76.7\dlt{$-$9.1} & 11.9\dlt{$-$76.0} \\
w/o lexical leg & 57.2\dlt{$+$0.2} & 85.2\dlt{$-$0.5} & 88.7\dlt{$+$0.7} \\
w/o tag leg & 55.0\dlt{$-$2.0} & -- & -- \\
\midrule
w/ consolidation & 55.6\dlt{$-$1.3} & 85.4\dlt{$-$0.3} & 86.9\dlt{$-$1.1} \\
w/ fact index & 55.2\dlt{$-$1.7} & 83.1\dlt{$-$2.6} & -- \\
\midrule
\memg{} & 55.3\dlt{$-$1.7} & 84.8\dlt{$-$0.9} & 86.0\dlt{$-$2.0} \\
\bottomrule
\end{tabular}
\end{table}

\textbf{Retrieval Legs.}
Removing the vector leg lowers the score by 35.7 and 76.0 points on OrgMemBench and LongMemEval$_\text{S}$ but only 9.1 on LoCoMo, where the lexical leg matches any query term rather than all of them (\Cref{tab:settings}).
Removing the lexical leg changes the score by at most 0.7 points.
The tag leg indexes only author labels and leaves the retrieved context byte-identical on the 73 OrgMemBench questions we compared, so its 2.0-point drop is run-to-run noise.

\textbf{Write-Time Processing.}
\mempp{} performs as well without consolidation or the fact index as with them, which shows that its accuracy does not depend on LLM calls at write time.
Indexing each original document whole also scores 2.6 points higher on LoCoMo than indexing extracted facts, even though the answerer reads the same original documents in both variants.
These results support the non-destructive design of \mempp{}, which keeps each document intact and leaves the choice between versions to read time.

\textbf{Entity Graph.}
\memg{} lowers the score by 0.9 and 2.0 points on LoCoMo and LongMemEval$_\text{S}$, while it led \mempp{} on LongMemEval$_\text{S}$ by at most 0.6 points with the other answerers.
Its gain is therefore not consistent across answerers.
On OrgMemBench, the answerer reads only the first 40,000 characters and never sees the graph triples, so this row does not measure the graph.

\section{Conclusion}
\label{sec:conclusion}
We have argued for a non-destructive view of organizational memory, where each document is kept whole rather than distilled at write time.
\mempp{} follows this view by keeping every version in one store and deferring the choice between versions to read time.
Experiments show that this design outperforms write-time extraction on OrgMemBench, achieves the best average LLM-judge score on LoCoMo and remains competitive on LongMemEval$_\text{S}$.
Ablations further show that semantic retrieval over original documents drives these results, whereas consolidation, fact extraction and an entity graph add cost without a consistent gain.
These findings suggest that keeping the original documents and choosing among them at read time is a strong baseline for organizational memory.
we acknowledge that few public benchmarks target organizational memory, so our organizational evaluation rests on the 73 questions that OrgMemBench asks about one synthetic organization.
In future work, we will collect more diverse organizational records and test \mempp{} on them to improve the system in this setting.

\clearpage

\subsection*{AI use statement}
In this work,the concept and architecture of \mempp{} are the authors' own work and design.
We used generative AI tools for partial help with generating the avatars using GPT-2, which we used later in our manually designed flowcharts and figures. We also utilized agentic AI tools to help cut the paper text to meet the page requirements. However, AI tools assisted the authors with setting the experiments and running scripts, three authors has been reviewing and monitoring the whole proposed memory system with result verifications over a period of roughly 4 months. We emphasize that the authors designed \mempp{} themselves, based on their review of prior work and the limitations they identified in existing systems.
Some parts of the text, such as the ablation study, were drafted by an LLM and then reviewed by the authors.
The cited works are valid systems that we compare with in the first place and well known. 
We have reviewed all AI-assisted work and take full responsibility for the content of this paper.

\subsection*{Reproducibility statement}

We will release the full system, including code and experiments will be open sourced for the community upon publication.

\bibliography{iclr2027_conference}

@article{kirchdorfer2026organizational,
  title={Organizational Memory for Agentic Business Process Execution},
  author={Kirchdorfer, Lukas and Rebmann, Adrian and Warmuth, Christian and Kampik, Timotheus and Heilker, Theiss and Berg, Gregor},
  journal={arXiv preprint arXiv:2607.03228},
  year={2026}
}

@article{zeng2025routine,
  title={Routine: A structural planning framework for llm agent system in enterprise},
  author={Zeng, Guancheng and Chen, Xueyi and Hu, Jiawang and Qi, Shaohua and Mao, Yaxuan and Wang, Zhantao and Nie, Yifan and Li, Shuang and Feng, Qiuyang and Qiu, Pengxu and others},
  journal={arXiv preprint arXiv:2507.14447},
  year={2025}
}

@misc{gardner2026orgmembench,
  title        = {OrgMemBench: A Benchmark for Long-Horizon Organizational Memory in AI Agents},
  author       = {Gardner, Jack},
  year         = {2026},
  note         = {Preprint},
  howpublished = {\url{https://github.com/JackCGardner/OrgMemBench}}
}

@article{walsh1991organizational,
  title={Organizational memory},
  author={Walsh, James P and Ungson, Gerardo Rivera},
  journal={Academy of management review},
  volume={16},
  number={1},
  pages={57--91},
  year={1991},
  publisher={Academy of Management Briarcliff Manor, NY 10510}
}

@article{chhikara2025mem0,
  title={Mem0: Building production-ready ai agents with scalable long-term memory},
  author={Chhikara, Prateek and Khant, Dev and Aryan, Saket and Singh, Taranjeet and Yadav, Deshraj},
  journal={arXiv preprint arXiv:2504.19413},
  year={2025}
}

@article{rasmussen2025zep,
  title={Zep: a temporal knowledge graph architecture for agent memory},
  author={Rasmussen, Preston and Paliychuk, Pavlo and Beauvais, Travis and Ryan, Jack and Chalef, Daniel},
  journal={arXiv preprint arXiv:2501.13956},
  year={2025}
}

@article{packer2023memgpt,
  title={Memgpt: Towards llms as operating systems},
  author={Packer, Charles and Wooders, Sarah and Lin, Kevin and Fang, Vivian and Patil, Shishir G and Stoica, Ion and Gonzalez, Joseph E},
  journal={arXiv preprint arXiv:2310.08560},
  year={2023}
}

@inproceedings{park2023generative,
  title={Generative agents: Interactive simulacra of human behavior},
  author={Park, Joon Sung and O'Brien, Joseph and Cai, Carrie Jun and Morris, Meredith Ringel and Liang, Percy and Bernstein, Michael S},
  booktitle={Proceedings of the 36th annual acm symposium on user interface software and technology},
  pages={1--22},
  year={2023}
}

@inproceedings{maharana2024evaluating,
  title={Evaluating very long-term conversational memory of llm agents},
  author={Maharana, Adyasha and Lee, Dong-Ho and Tulyakov, Sergey and Bansal, Mohit and Barbieri, Francesco and Fang, Yuwei},
  booktitle={Proceedings of the 62nd Annual Meeting of the Association for Computational Linguistics (Volume 1: Long Papers)},
  pages={13851--13870},
  year={2024}
}

@article{wu2024longmemeval,
  title={Longmemeval: Benchmarking chat assistants on long-term interactive memory},
  author={Wu, Di and Wang, Hongwei and Yu, Wenhao and Zhang, Yuwei and Chang, Kai-Wei and Yu, Dong},
  journal={arXiv preprint arXiv:2410.10813},
  year={2024}
}

@article{gutierrez2024hipporag,
  title={Hipporag: Neurobiologically inspired long-term memory for large language models},
  author={Guti{\'e}rrez, Bernal J and Shu, Yiheng and Gu, Yu and Yasunaga, Michihiro and Su, Yu},
  journal={Advances in neural information processing systems},
  volume={37},
  pages={59532--59569},
  year={2024}
}

@article{edge2024local,
  title={From local to global: A graph rag approach to query-focused summarization},
  author={Edge, Darren and Trinh, Ha and Cheng, Newman and Bradley, Joshua and Chao, Alex and Mody, Apurva and Truitt, Steven and Metropolitansky, Dasha and Ness, Robert Osazuwa and Larson, Jonathan},
  journal={arXiv preprint arXiv:2404.16130},
  year={2024}
}

@article{lewis2020retrieval,
  title={Retrieval-augmented generation for knowledge-intensive nlp tasks},
  author={Lewis, Patrick and Perez, Ethan and Piktus, Aleksandra and Petroni, Fabio and Karpukhin, Vladimir and Goyal, Naman and K{\"u}ttler, Heinrich and Lewis, Mike and Yih, Wen-tau and Rockt{\"a}schel, Tim and others},
  journal={Advances in neural information processing systems},
  volume={33},
  pages={9459--9474},
  year={2020}
}

@inproceedings{karpukhin2020dense,
  title={Dense passage retrieval for open-domain question answering},
  author={Karpukhin, Vladimir and Oguz, Barlas and Min, Sewon and Lewis, Patrick and Wu, Ledell and Edunov, Sergey and Chen, Danqi and Yih, Wen-tau},
  booktitle={Proceedings of the 2020 conference on empirical methods in natural language processing (EMNLP)},
  pages={6769--6781},
  year={2020}
}

@article{xu2026mem,
  title={A-mem: Agentic memory for llm agents},
  author={Xu, Wujiang and Liang, Zujie and Mei, Kai and Gao, Hang and Tan, Juntao and Zhang, Yongfeng},
  journal={Advances in Neural Information Processing Systems},
  volume={38},
  pages={17577--17604},
  year={2026}
}

@article{nan2025nemori,
  title={Nemori: Self-organizing agent memory inspired by cognitive science},
  author={Nan, Jiayan and Ma, Wenquan and Wu, Wenlong and Chen, Yize},
  journal={arXiv e-prints},
  pages={arXiv--2508},
  year={2025}
}

@inproceedings{fang2026lightmem,
  title={Lightmem: Lightweight and efficient memory-augmented generation},
  author={Fang, Jizhan and Deng, Xinle and Xu, Haoming and Jiang, Ziyan and Tang, Yuqi and Xu, Ziwen and Deng, Shumin and Yao, Yunzhi and Wang, Mengru and Qiao, Shuofei and others},
  booktitle={International Conference on Learning Representations},
  volume={2026},
  pages={98706--98729},
  year={2026}
}

@inproceedings{huang2026licomemory,
  title={Licomemory: Lightweight and cognitive agentic memory for efficient long-term reasoning},
  author={Huang, Zhengjun and Tian, Zhoujin and Guo, Qintian and Zhang, Fangyuan and Zhou, Yingli and Jiang, Di and Xie, Zeying and Zhou, Xiaofang},
  booktitle={Findings of the Association for Computational Linguistics: ACL 2026},
  pages={36842--36858},
  year={2026}
}

@article{zhang2026gmemory,
  title={G-memory: Tracing hierarchical memory for multi-agent systems},
  author={Zhang, Guibin and Fu, Muxin and Wang, Kun and Wan, Frank and Yu, Miao and Yan, Shuicheng},
  journal={Advances in Neural Information Processing Systems},
  volume={38},
  pages={12988--13018},
  year={2026}
}

@article{kim2025premem,
  title={Pre-storage reasoning for episodic memory: Shifting inference burden to memory for personalized dialogue},
  author={Kim, Sangyeop and Lee, Yohan and Kim, Sanghwa and Kim, Hyunjong and Cho, Sungzoon},
  journal={arXiv preprint arXiv:2509.10852},
  year={2025}
}

@article{li2025memos,
  title={Memos: A memory os for ai system},
  author={Li, Zhiyu and Xi, Chenyang and Li, Chunyu and Chen, Ding and Chen, Boyu and Song, Shichao and Niu, Simin and Wang, Hanyu and Yang, Jiawei and Tang, Chen and others},
  journal={arXiv preprint arXiv:2507.03724},
  year={2025}
}

@inproceedings{zhong2024memorybank,
  title={Memorybank: Enhancing large language models with long-term memory},
  author={Zhong, Wanjun and Guo, Lianghong and Gao, Qiqi and Ye, He and Wang, Yanlin},
  booktitle={Proceedings of the AAAI conference on artificial intelligence},
  volume={38},
  number={17},
  pages={19724--19731},
  year={2024}
}

@inproceedings{cormack2009reciprocal,
  title={Reciprocal rank fusion outperforms condorcet and individual rank learning methods},
  author={Cormack, Gordon V and Clarke, Charles LA and Buettcher, Stefan},
  booktitle={Proceedings of the 32nd international ACM SIGIR conference on Research and development in information retrieval},
  pages={758--759},
  year={2009}
}

@misc{langmem2025,
  author       = {{LangChain AI}},
  title        = {{LangMem}: Long-Term Memory {SDK} for {LLM} Agents},
  howpublished = {\url{https://langchain-ai.github.io/langmem/}},
  year         = {2025},
  note         = {Accessed: 2025-11-20}
}
\bibliographystyle{iclr2027_conference}

\newpage
\appendix
\begin{center}
{\LARGE\bfseries \mempp{} Appendix}
\end{center}

\section{Method Details}
\label{app:method}
\Cref{alg:mempp} summarizes the write and read paths of \mempp{}, and \Cref{alg:consolidation} summarizes the optional consolidation operator.
Here $f_{\mathrm{LLM}}$ is a language model, $f_{\mathrm{ans}}$ is the answering model and $\mathcal{K}$ is the entity graph of \memg{}.
Each $\mathcal{P}$ is a fixed prompt, $\|$ denotes concatenation and $\tau$ is the similarity threshold of consolidation.

\begin{algorithm}[h]
\caption{\mempp{} Write and Read Paths}
\label{alg:mempp}
\begin{algorithmic}[1]
\Require Document stream $\mathcal{D}$, question $q$, optional as-of date $\theta$
\Ensure Store $\mathcal{R}$, evidence $E$, answer $\hat{y}$
\Statex \textit{\% --- Write path (\S\ref{subsec:write}) ---}
\For{each document $(x, a, t) \in \mathcal{D}$}
  \If{$x$ is empty or exactly repeats an active row in the author scope of $a$}
    \State \textbf{continue} \Comment{no row is added}
  \EndIf
  \State $\mathbf{e} \gets h(x)$ \Comment{no generative call}
  \State $r \gets \langle x, a, t, \mathbf{e}, \sigma \rangle$ with $\sigma$ set to active and current
  \State $\mathcal{R} \gets \mathcal{R} \cup \{r\}$
  \State Index $x$ for lexical search, $a$ as a tag and $\mathbf{e}$ for vector search
  \If{running \memg{}}
    \State $\mathcal{K} \gets \mathcal{K} \cup f_{\mathrm{LLM}}(\mathcal{P}_{\mathrm{triple}} \,\|\, x)$ \Comment{two LLM calls per document or chunk}
  \EndIf
\EndFor
\If{consolidation is enabled}
  \State $\mathcal{C}(\mathcal{R}, \tau)$ \Comment{\Cref{alg:consolidation}, off in headline runs}
\EndIf
\Statex \textit{\% --- Read path (\S\ref{subsec:read}) ---}
\State $\mathcal{S}_\theta \gets \{ r \in \mathcal{R} \mid \mathrm{active}(\sigma_r),\ t_r \le \theta \}$ \Comment{time condition only if $\theta$ is given}
\For{$\ell \in \{\mathrm{lex}, \mathrm{tag}, \mathrm{vec}\}$}
  \State $L^{\theta}_\ell \gets$ at most $C$ rows of $\mathcal{S}_\theta$ ranked by index $\ell$ for $q$
\EndFor
\State Score each $r \in \bigcup_\ell L^{\theta}_\ell$ by $\mathrm{RRF}(r) = \sum_{\ell:\, r \in L^{\theta}_\ell} w_\ell / (\gamma + \mathrm{rank}_\ell(r))$
\State $E \gets$ the first $k-3$ rows by score, with ties broken by ingest recency
\State $E \gets E \,\|\,$ the three latest-dated remaining matched rows \Comment{undated rows use ingest time}
\State Fill any missing reserve slot with the most recently ingested unmatched rows
\If{running \memg{}}
  \State $V_q \gets f_{\mathrm{LLM}}(\mathcal{P}_{\mathrm{entity}} \,\|\, q)$ \Comment{entities named in $q$}
  \State $E \gets E \,\|\,$ at most five triples of $\mathcal{K}$ reached from $V_q$ \Comment{no as-of filter}
\EndIf
\State $\hat{y} \gets f_{\mathrm{ans}}(q \,\|\, E)$ with rows in rank order, each preceded by its date and author where recorded
\end{algorithmic}
\end{algorithm}

\begin{algorithm}[t]
\caption{Consolidation Operator $\mathcal{C}$}
\label{alg:consolidation}
\begin{algorithmic}[1]
\Require Store $\mathcal{R}$, similarity threshold $\tau$
\Ensure Updated states $\sigma$ and merged rows
\State Group current rows greedily by cosine $\ge \tau$ to the first row of each group
\For{each group $B$ with more than one row}
  \State $y(B) \gets f_{\mathrm{LLM}}(\mathcal{P}_{\mathrm{label}} \,\|\, B)$ \Comment{skip $B$ if the call fails}
  \If{$y(B) = \textsc{conflict}$}
    \State $r^\ast \gets$ the row of $B$ with the latest event date
    \State Mark every other row of $B$ as superseded by $r^\ast$ in $\sigma$ \Comment{row stays active}
  \ElsIf{$y(B) = \textsc{restatement}$}
    \State $x' \gets f_{\mathrm{LLM}}(\mathcal{P}_{\mathrm{merge}} \,\|\, B)$ \Comment{at most 220 tokens}
    \State Add and index the active row $\langle x', A_B, \min_{r \in B} t_r, h(x'), \sigma \rangle$ \Comment{$A_B$ is every author of $B$}
    \State Mark the rows of $B$ as archived in $\sigma$ \Comment{removed from retrieval}
  \Else
    \State Leave $B$ untouched \Comment{\textsc{distinct}}
  \EndIf
\EndFor
\end{algorithmic}
\end{algorithm}

\paragraph{Consolidation Operator ($\mathcal{C}$).}
Unlike memories that run a language model on every incoming document~\citep{chhikara2025mem0,rasmussen2025zep,xu2026mem}, $\mathcal{C}$ runs between writes and queries over documents already stored.
It spends a model call only where two documents likely describe the same fact.
The operator first greedily groups current rows, where each group takes the rows whose embeddings have a cosine similarity of at least $\tau$ to its first row.
It then asks a language model what changed within each group $B$ and assigns the group a label
\begin{equation}
  y(B) \in \{\textsc{conflict},\ \textsc{restatement},\ \textsc{distinct}\} ,
\end{equation}
where a conflict means one fact took a new value over time, a restatement means the rows say the same thing in different words, and distinct means the rows only look alike.
In a conflict, the row with the latest event date stays current.
Each older row keeps its text and date and is marked in $\sigma$ as superseded by that row, and it therefore remains retrievable with its own date.
In a restatement, a second model call writes one merged row of at most 220 tokens.
This row takes the earliest date in $B$ and lists every original author.
The originals are marked as archived in $\sigma$, which removes them from retrieval.
These merged rows are the only rows in the store whose text a model has written.
Because of this earliest date, a merged row can carry text from later rows past the as-of date $\theta$.
This affects only the consolidation ablation, since consolidation is off by default.
A distinct group is left untouched, and so is any group whose labeling call fails.
The cost is therefore at most two model calls per group rather than one per document.

\section{Additional Experimental Details}
\label{app:exp}

\subsection{Baselines}
\label{app:baselines}

\textbf{Full Context} places the whole history in the prompt of the answering model and uses no external memory.

\textbf{RAG} \citep{lewis2020retrieval,karpukhin2020dense} splits the history into fixed-size chunks and returns the chunks whose embeddings are closest to the question.
On LoCoMo, the suffix gives the chunk size in tokens.
Unlike \mempp{}, it can split one document across several chunks.

\textbf{Mem0} \citep{chhikara2025mem0} uses an LLM to extract salient facts from each new message and compares each fact with the most similar stored memories.
An LLM then adds, updates or deletes a memory or leaves the store unchanged, so a newer fact can overwrite an older one.

\textbf{Zep} \citep{rasmussen2025zep} builds a temporal knowledge graph with its Graphiti engine.
It extracts entities and facts from each episode and stores each fact as an edge with the period over which it held.
When a later fact contradicts an edge, Zep invalidates that edge.
At query time it combines semantic, keyword and graph search over facts, entities and communities.

\textbf{A-Mem} \citep{xu2026mem} follows the Zettelkasten method and stores each interaction as a note with keywords, tags and a context description written by an LLM.
It links each new note to related notes and can update the attributes of older notes as new ones arrive.

\textbf{Nemori} \citep{nan2025nemori} segments a conversation into coherent episodes and rewrites each episode as a narrative.
It then distills semantic knowledge by predicting a new episode from existing memory and storing what the prediction missed.

\textbf{gbrain} \citep{gardner2026orgmembench} is an open-source memory system that combines keyword, vector and graph search over a store extracted from its documents.
We report it only on OrgMemBench, whose paper also evaluates it.

\subsection{Evaluation Metrics}
\label{app:metrics}

\textbf{LLM-Judge Score.}
Our primary metric asks gpt-4o-mini to judge each generated answer against the gold answer.
On LoCoMo and LongMemEval$_\text{S}$ the judge labels each answer as correct or wrong.
The score is the share of correct answers,
\begin{equation}
  \mathrm{LLM} = \frac{100}{N} \sum_{i=1}^{N} \mathbf{1}\big[J(q_i, a_i, \hat{a}_i) = \textsc{correct}\big] ,
\end{equation}
where $J$ is the judge and $N$ is the number of questions $q_i$ with gold answers $a_i$ and generated answers $\hat{a}_i$.
LoCoMo uses the Mem0 judge prompt \citep{chhikara2025mem0} over 1,520 questions.
On LongMemEval$_\text{S}$ we use one prompt per question type following Zep \citep{rasmussen2025zep} and score our runs on its 470 non-abstention questions.
Published baseline scores on LongMemEval$_\text{S}$ cover all 500 questions.
On OrgMemBench the 73 questions carry rubrics with 247 weighted facets in total.
In place of the default judge of the benchmark, gpt-4o-mini grades each answer against every facet of its rubric at temperature zero.
This gives a question score $s_i \in [0,1]$ that rewards partial coverage, and the overall score is $\frac{100}{N}\sum_{i} s_i$.
We judge each of our runs three times and report the mean.

\textbf{F1 Score.}
F1 is the harmonic mean of token-level precision $P$ and recall $R$ between the generated and gold answers after standard answer normalization,
\begin{equation}
  \mathrm{F1} = \frac{2PR}{P+R} , \qquad
  P = \frac{|\hat{T} \cap T|}{|\hat{T}|} , \qquad
  R = \frac{|\hat{T} \cap T|}{|T|} ,
\end{equation}
where $\hat{T}$ and $T$ are the token multisets of the generated and gold answers.

\textbf{BLEU-1.}
BLEU-1 is the unigram precision of the generated answer against the gold answer, scaled by a brevity penalty,
\begin{equation}
  \text{BLEU-1} = \mathrm{BP} \cdot \frac{\sum_{w} \min\big(c_{\hat{a}}(w),\, c_{a}(w)\big)}{\sum_{w} c_{\hat{a}}(w)} , \qquad
  \mathrm{BP} =
  \begin{cases}
    1 & \text{if } |\hat{a}| > |a| , \\
    e^{\,1 - |a|/|\hat{a}|} & \text{otherwise} ,
  \end{cases}
\end{equation}
where $c_{x}(w)$ counts word $w$ in answer $x$ and $|x|$ is the length of $x$ in words.
We report F1 and BLEU-1 on LoCoMo only.
Both reward exact wording and can penalize a correct answer phrased differently, which is why the LLM-judge score is our primary metric.

\subsection{Implementation Details}
\label{app:impl}
Table~\ref{tab:settings} lists the retrieval settings that differ across benchmarks.
All runs share the fusion constant $\gamma = 60$, $k = 50$ retrieved rows and three recency slots.
A disjunctive lexical index returns rows that match any query term, whereas a conjunctive one returns only rows that contain every term.
On OrgMemBench the encoder runs locally and embeds at most 256 word pieces of each document, while the stored text and the lexical index keep the full document.
Only the 15 Audit Replay questions of OrgMemBench carry an as-of date, so the date bound applies only to them.
On OrgMemBench the benchmark passes only the first 40,000 characters of the retrieved context to the answering model, and these hold about 19 of the 50 rows.
The recency slots and the \memg{} triples come after the fused rows, so they never reach the answering model on this benchmark.
\memg{} calls gpt-4o-mini twice per document at write and once per question at read.
On the conversational benchmarks, its write calls run per ten-turn chunk instead of per document.

\begin{table}[t]
\centering
\caption{Category distributions of the OrgMemBench~\citep{gardner2026orgmembench}, LoCoMo~\citep{maharana2024evaluating} and LongMemEval\textsubscript{S}~\citep{wu2024longmemeval} datasets.}
\label{tab:datasets}
\small
\setlength{\tabcolsep}{4pt}
\begin{tabular}{@{}lr lr lr@{}}
\hline\noalign{\vskip 2pt}
\multicolumn{2}{c}{\textbf{OrgMemBench} (medium)} & \multicolumn{2}{c}{\textbf{LoCoMo}} & \multicolumn{2}{c}{\textbf{LongMemEval\textsubscript{S}}} \\
\cline{1-2}\cline{3-4}\cline{5-6}\noalign{\vskip 2pt}
Category & Count & Category & Count & Category & Count \\
\noalign{\vskip 2pt}\hline\noalign{\vskip 2pt}
C1 Supersession  & 15 & Single-hop   &  841 & Knowledge update          &  72 \\
C2 Provenance    & 15 & Multi-hop    &  282 & Multi-session             & 121 \\
C3 Bi-temporal   &  7 & Temporal     &  321 & Temporal reasoning        & 127 \\
C4 Audit replay  & 15 & Open domain  &   96 & Single-session user       &  64 \\
C5 Justification & 15 & --           &   -- & Single-session assistant  &  56 \\
C6 Contradiction &  6 & --           &   -- & Single-session preference &  30 \\
\noalign{\vskip 2pt}\hline\noalign{\vskip 2pt}
\textbf{Total} & \textbf{73} & \textbf{Total} & \textbf{1,540} & \textbf{Total} & \textbf{470} \\
\noalign{\vskip 2pt}\hline
\end{tabular}
\end{table}

\begin{table}[t]
\centering
\caption{Retrieval settings that differ across benchmarks.}
\label{tab:settings}
\small
\setlength{\tabcolsep}{4pt}
\begin{tabular}{lccc}
\toprule
Setting & OrgMemBench & LoCoMo & LongMemEval$_\text{S}$ \\
\midrule
Embedding model & all-MiniLM-L6-v2 & text-embedding-3-small & text-embedding-3-small \\
Embedding size & 384 & 1536 & 1536 \\
Weights $(w_\text{lex}, w_\text{tag}, w_\text{vec})$ & (1, 1, 4) & (1, 1, 2) & (1, 1, 4) \\
Candidates per index $C$ & 50 & 200 & 50 \\
Lexical matching & conjunctive & disjunctive & conjunctive \\
\bottomrule
\end{tabular}
\end{table}

\subsection{Sensitivity to the Number of Retrieved Rows}
\label{app:topk}
\begin{figure}[H]
\centering
\includegraphics[width=0.8\linewidth]{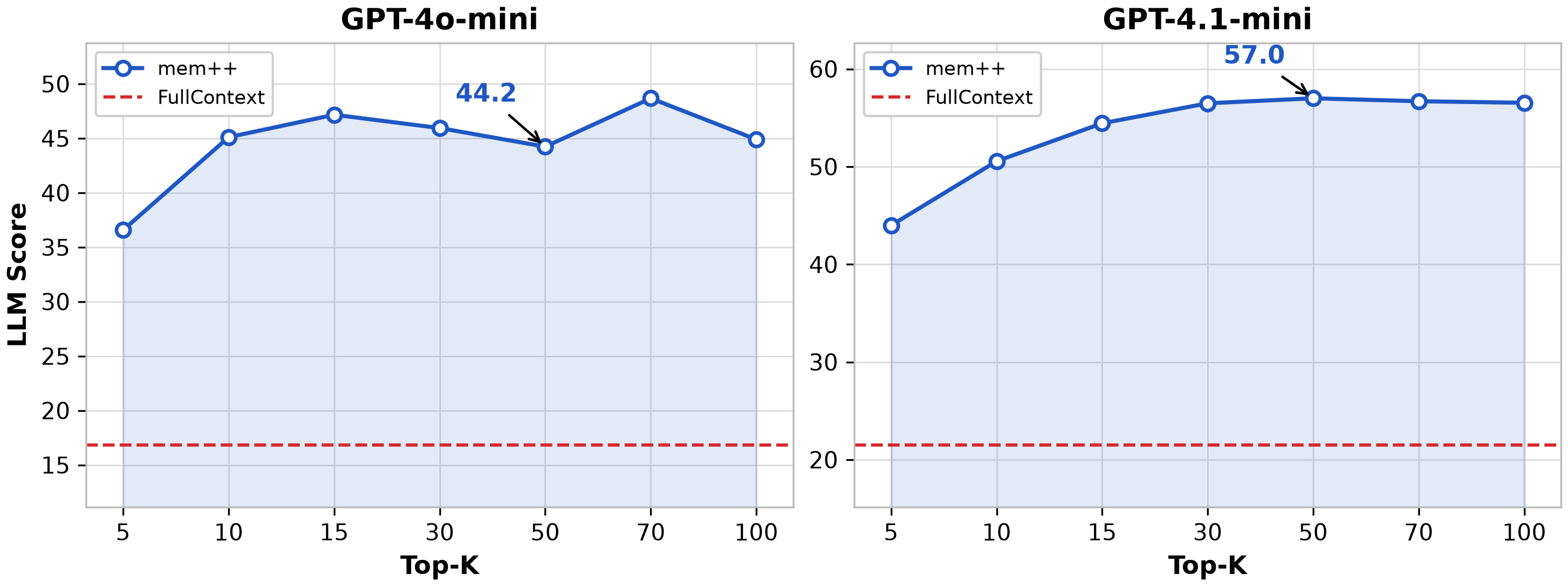}
\caption{OrgMemBench score of \mempp{} as the number of retrieved rows $k$ varies.
The dashed line is Full Context, and the dotted line marks $k=50$ used in the main experiments.}
\label{fig:topk}
\end{figure}

\Cref{fig:topk} and \Cref{tab:topk} vary the number of retrieved rows $k$ on OrgMemBench with both answering models, while all other settings follow \Cref{app:impl}.
With gpt-4.1-mini, the score rises from 40.5 at $k=5$ to 56.5 at $k=30$ and then stays between 56.5 and 57.0 up to $k=100$.
With gpt-4o-mini, it rises from 36.6 at $k=5$ to between 44.2 and 48.7 for every $k \ge 10$, with no clear trend beyond $k=15$.
The plateau is partly mechanical, because the benchmark passes only the first 40,000 characters of the context to the answering model, which hold about 19 rows and at most 25 (\Cref{app:impl}).
For $k \ge 30$ the fused rows alone exceed this budget, and their order does not depend on $k$ since each index returns at most $C$ candidates.
The answering model therefore reads the same rows for every $k \ge 30$, and the changes across these settings are run-to-run variation, which reaches 4.4 points with gpt-4o-mini.

\begin{table}[H]
\centering
\caption{OrgMemBench score of \mempp{} for each $k$.
Retrieved tokens are summed over all questions and counted before the 40,000-character cut.}
\label{tab:topk}
\small
\begin{tabular}{lccc}
\toprule
$k$ & gpt-4o-mini & gpt-4.1-mini & Retrieved tokens \\
\midrule
5   & 36.57 & 43.99 & 0.20M \\
10  & 45.10 & 50.56 & 0.41M \\
15  & 47.16 & 54.44 & 0.61M \\
30  & 45.94 & 56.49 & 1.22M \\
50  & 44.23 & 57.00 & 2.01M \\
70  & 48.67 & 56.70 & 2.77M \\
100 & 44.89 &56.54 & 3.63M \\
\midrule
Full Context & 16.85 & 21.52 & 18.68M \\
\bottomrule
\end{tabular}
\end{table}

\end{document}